# SEB-ChOA: An Improved Chimp Optimization Algorithm Using Spiral Exploitation Behavior

Leren Qian, Mohammad Khishe, Yiqian Huang, Seyedali Mirjalili

*Abstract*

The chimp optimization algorithm (ChOA) is a nature-inspired algorithm that imitates chimpanzees' individual intelligence and hunting behaviors. In this algorithm, the hunting process consists of four steps: driving, blocking, chasing, and attacking. Because of the novelty of ChOA, the steps of the hunting process have been modeled in the simplest possible way, leading to slow and premature convergence similar to other iterative algorithms. This paper proposes six spiral functions and introduces two novel hybrid spiral functions (SEB-ChOA) to rectify the abovementioned deficiencies. The SEB-ChOAs' performance is evaluated on 23 standard benchmarks, 20 benchmarks of IEEE CEC-2005, 10 cases of IEEE CEC06-2019 test-suite, and 12 constrained real-world engineering problems of IEEE CEC-2020. The SEB-ChOAs are compared with three groups of optimization algorithms, including Particle Swarm Optimization (PSO) and Genetic Algorithm (GA) as the most well-known optimization algorithms, Slime Mould Algorithm (SMA), Marine Predators Algorithm (MPA), Ant Lion Optimization (ALO), Henry Gas Solubility Optimization (HGSO), as almost novel optimization algorithms, and jDE100 and DISHchain1e+12, as winners of IEEE CEC06-2019 competition, and also EBOwithCMAR and CIPDE as superior secondary optimization algorithms. The SEB-ChOAs reached the first rank among almost all benchmarks and demonstrated very competitive results compared to jDE100 and DISHchain1e+12 as the best-performing optimizers. Statistical evidence shows that the SEB-ChOA outperforms the PSO, GA, SMA, MPA, ALO, and HGSO optimizers while producing results comparable to those of the jDE100 and DISHchain1e+12 algorithms.

**Keywords**: chimp optimization algorithm; spiral model; metaheuristics; global optimization; constrained engineering design

## 1 Introduction

Spiral functions have proved effective in a variety of optimization problems because their geometric structure allows fine control over search direction and step size. In metaheuristic algorithms, spiral motions can be used to generate new candidate solutions with desirable properties such as global search ability, diversity preservation, reliability, and favorable convergence characteristics. By adjusting the slope and amplitude of a spiral trajectory, one can tune the balance between fast convergence and thorough exploitation of the search space.

The convergence speed, final accuracy, and ability to escape local minima in a metaheuristic are strongly influenced by the trajectory and amplitude of the underlying movement model. Steep spiral trajectories can accelerate convergence and exploration, whereas shallow spirals promote more local exploitation. Properly chosen spiral dynamics can therefore provide a better trade-off between exploration and exploitation than purely random walks or simple linear movements.

The chimp optimization algorithm (ChOA) is a recent metaheuristic that models the cooperative hunting behavior of chimpanzees. In ChOA, chimps are divided into four functional roles—attacker, driver, chaser, and barrier—and move according to simple position-update equations. While ChOA has shown promising results, its exploitation phase is modeled by chaotic maps in a rather simplistic way, which may lead to divergence at the final stage and a low overall convergence rate.

In this paper we introduce spiral exploitation behavior (SEB) into ChOA. We consider six well-known spiral functions and design two new hybrid spirals with larger amplitudes in early iterations and steeper downward slopes in later iterations. These shapes aim to enhance exploration in the beginning of the search and accelerate exploitation near convergence. Extensive experiments on classical, competition, and real-world benchmark suites are conducted to study how different spiral models affect the algorithm's behavior and performance.

## 2 Related Work

Many engineering applications now rely on metaheuristic optimization, including vehicle routing and scheduling, resource allocation, data management, defense and aerospace, social networks, soil and civil engineering, wireless sensor networks, energy systems, and pattern classification. Spiral shapes have recently attracted particular attention as a way to improve the search pattern of these algorithms.

Existing studies have used spiral motions in different ways: as stand-alone spiral optimizers, as spiral-based movement strategies in PSO, as hybrid components combined with artificial bee colony or whale optimization algorithms, and as enhancements to differential evolution and Harris Hawks optimization. Other works studied spiral-based initialization of particle swarms or combined spiral search patterns with chaos theory to address non-convex global optimization problems.

Overall, these studies reveal that spiral-based movement can improve global exploration, help avoid premature stagnation, and lead to better convergence on constrained and non-convex problems. However, the literature still has several gaps: (i) many methods have been tested only on simple or low-dimensional problems; (ii) there is no systematic framework for integrating spiral shapes into different metaheuristics; and (iii) comprehensive empirical comparisons between different spiral models are scarce. This motivates a more systematic study of spiral exploitation behavior and its hybridization with an existing algorithm such as ChOA.

## 3 Chimp Optimization Algorithm (ChOA)

ChOA is a population-based metaheuristic inspired by chimp group hunting. The population consists of chimpanzees whose positions represent candidate solutions. At each iteration, chimps move according to the positions of the four best individuals, which are assigned the roles of attacker, barrier, chaser, and driver.

Let $t$ be the iteration index, $x_{\text{chimp}}(t)$ the position of a chimp, and $x_{\text{prey}}(t)$ the best solution found so far (prey). The basic update is

$$d = c \cdot x_{\text{prey}}(t) - m \cdot x_{\text{chimp}}(t),$$
$$x_{\text{chimp}}(t+1) = x_{\text{prey}}(t) - a \cdot d,$$

where $a$, $c$, and $m$ are coefficient vectors determined by chaotic maps and random factors. A nonlinear control vector $f$ decreases from 2.5 to 0 to gradually reduce the step size and encourage convergence.

The social roles are modeled by four distance terms $d_{\text{attacker}}, d_{\text{barrier}}, d_{\text{chaser}}, d_{\text{driver}}$ and corresponding candidate positions $x_1, x_2, x_3, x_4$. The new position of each chimp is computed as the average of these four role-based positions.

In the original ChOA, the exploitation phase—when chimps close in on the prey—is modeled using a simple chaotic-vector replacement with probability 0.5. While this introduces randomness, it does not reflect realistic spiral-like trajectories observed in actual hunting and can degrade convergence quality. This motivates replacing the chaotic exploitation behavior with more structured spiral dynamics.

## 4 SEB-ChOA: Spiral exploitation behavior

### 4.1 Spiral exploitation model

To improve exploitation, we modify the attacking behavior of ChOA by introducing spiral movement around the prey. When a random parameter $\lambda$ exceeds 0.5, chimps follow a spiral trajectory toward the prey; otherwise, they use the standard ChOA update. Formally, the new position is

$$x(t+1) = \begin{cases} x_{\text{prey}}(t) - a \cdot d, & \lambda < 0.5, \\ D_0 \cdot M \cdot \cos(2\pi t) + x_{\text{prey}}(t), & \lambda \geq 0.5, \end{cases}$$

where $D_0 = x_{\text{prey}}(t) - x(t)$ is the distance to the prey and $M$ is a spiral shape term. By controlling $M$, different spiral trajectories are obtained.

### 4.2 Canonical spiral shapes

We consider six canonical spiral models:

- **Archimedean spiral**: $r = a\theta$, where the distance between successive turns is constant.
- **Logarithmic spiral**: $\log r = a\theta$, producing a constant angle between radius and tangent.
- **Fermat spiral**: $r^2 = a^2\theta$, with quadratic radial growth.
- **Lituus spiral**: $r^2 = a^2/\theta$, the inverse of Fermat's spiral.
- **Equiangular spiral**: $\ln r = a\theta$, closely related to the logarithmic spiral.
- **Random spiral**: $r = \text{rand}() \cdot \theta$, where the slope parameter is randomly sampled at each step.

Here $r$ is the radius, $\theta$ the polar angle, and $a$ a constant controlling the slope.

### 4.3 Hybrid spiral shapes

To further enhance performance, two new **hybrid spiral shapes** are proposed:

- **Hybrid Spiral Shape 1 (HSS1)**:

$$r \cdot \log r = a\theta,$$

which yields large amplitudes in early iterations and a moderate downward slope later, helping exploration first and exploitation later.

- **Hybrid Spiral Shape 2 (HSS2)**:

$$r^2 \cdot \log r = a\theta,$$

which has even larger initial amplitudes and a steeper downward slope, emphasizing aggressive exploration followed by fast convergence.

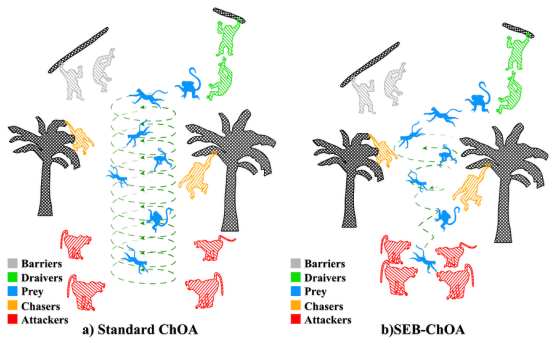

Fig.1. spiral mechanism illustration

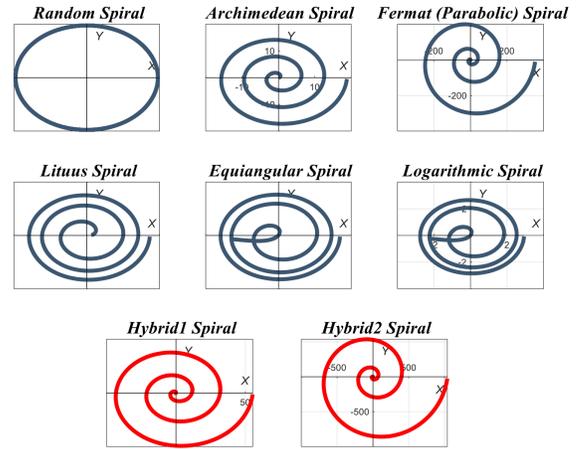

Fig. 2. 2D spiral comparison

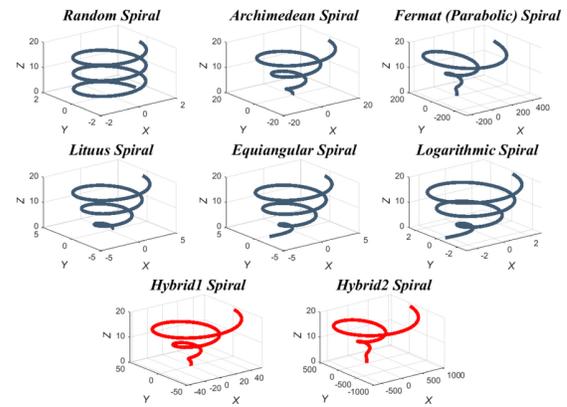

Fig. 3. 3D spiral comparison

Each spiral is integrated into the ChOA exploitation step, producing eight SEB-ChOA variants (six canonical, two hybrid). In all cases, the rest of the ChOA framework—including role assignment, coefficient-vector updates, and population management—remains unchanged.

## 5 Simulation Results

### 5.1 Benchmark sets

SEB-ChOA is evaluated on four groups of benchmarks:

1. **Standard functions**: 23 widely used functions including unimodal, multimodal, and fixed-

dimension multimodal problems. This group is used to study convergence speed, exploitation quality, and robustness to local optima.
2. **CEC 2005 real-parameter suite**: 20 transformed problems (shifted, rotated, expanded or combined) grouped into unimodal, multimodal, and composite categories.
3. **CEC06-2019 100-Digit Challenge**: 10 challenging functions where each correct digit in the objective value earns one point up to 100.
4. **CEC 2020 real-world constrained problems**: 12 real-world engineering design tasks from chemical process design, mechanical structures, electrical systems, power electronics, and livestock feed optimization.

Baseline algorithms include PSO, GA, SMA, MPA, ALO, HGSO, jDE100, DISHchain1e−12, EBOwithCMAR, CIPDE, and the original ChOA. All methods are implemented in MATLAB and run 30 independent times on each problem. Performance is measured by average and standard deviation of the objective value, and statistical significance is tested using the Wilcoxon rank-sum test at a 5% significance level.

For visualization, convergence curves and performance comparisons are summarized in the following figures:

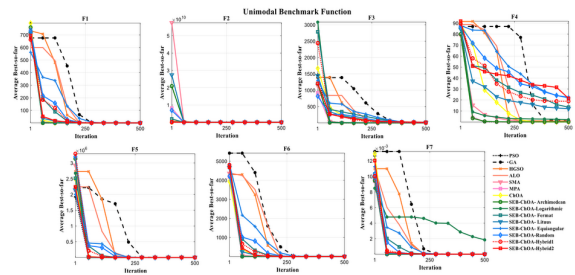

Fig. 4. unimodal convergence

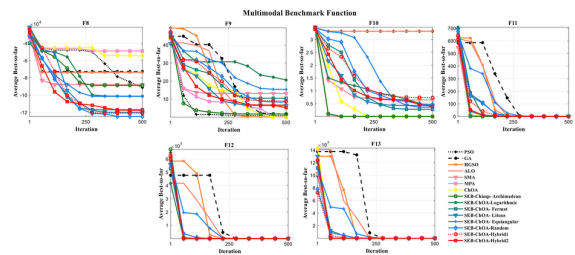

Fig. 5. multimodal convergence

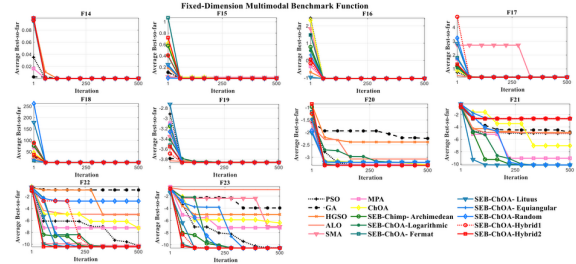

Fig. 6. fixed-dimension multimodal

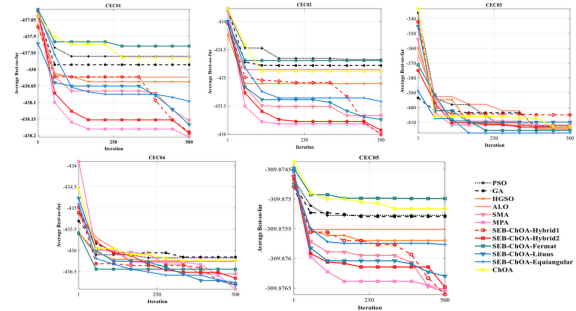

Fig. 7. CEC05 unimodal

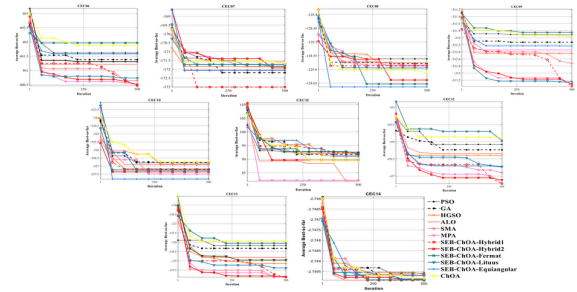

Fig. 8. CEC05 multimodal

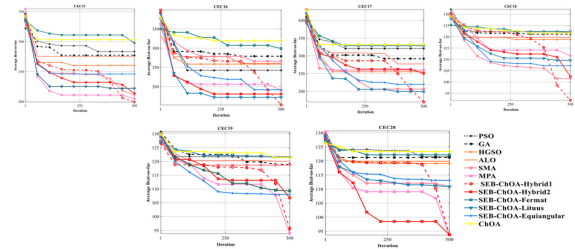

Fig. 9. real-world constrained problems

## 5.2 Unimodal benchmarks

Unimodal functions with a single global optimum are used to assess exploitation and convergence speed.

Across these problems, SEB-ChOA with HSS1 achieves the best average ranking and the fastest convergence in most cases. HSS2 is usually the second best, with slightly more aggressive behavior that can occasionally lead to overshooting on certain functions.

The convergence curves in **Fig. 4** show that both HSS1 and HSS2 reduce the objective value much faster than the canonical spiral variants and the original ChOA, especially in early and mid iterations. Wilcoxon tests confirm that HSS1 significantly outperforms PSO, GA, SMA, MPA, ALO, HGSO, and standard ChOA on the majority of unimodal functions.

### 5.3 Multimodal and fixed-dimension multimodal benchmarks

Multimodal problems with many local optima are used to evaluate exploration ability and robustness against premature convergence. Fixed-dimension multimodal functions further remove the confounding effect of changing dimensionality.

On these problems, SEB-ChOAs maintain strong performance. The hybrid variants, particularly HSS1, consistently achieve lower average objective values and smaller standard deviations than the baselines. The convergence plots in **Figs. 5–6** indicate that spiral exploitation behavior allows SEB-ChOA to escape local minima more effectively while still converging quickly to high-quality solutions.

### 5.4 CEC 2005 and CEC06-2019 benchmark suites

On the CEC 2005 problems, SEB-ChOA-HSS1 again attains the best overall ranking, with SEB-ChOA-HSS2 and the best canonical spirals close behind. The method produces particularly strong results on composite functions that combine multiple shifted and rotated components.

In the CEC06-2019 100-Digit Challenge, SEB-ChOAs achieve highly competitive scores relative to competition winners jDE100 and DISHchain1e−12. Although the latter were specifically designed for this suite, HSS1 remains comparable or superior on several problems, indicating that spiral exploitation can generalize beyond simple function landscapes.

### 5.5 Real-world constrained engineering problems

The real-world constrained problems from CEC 2020 pose complex objective landscapes with multiple nonlinear constraints. Here SEB-ChOAs are integrated with a standard constraint-handling mechanism and evaluated on 12 engineering tasks.

The results summarized in **Fig. 9** show that HSS1 outperforms the original ChOA and most competitor algorithms on the majority of problems, often achieving lower objective values while satisfying all constraints. This demonstrates that spiral exploitation behavior is not limited to artificial test functions and can also benefit real engineering optimization tasks.

### 6  Conclusion

This paper introduced SEB-ChOA, a spiral exploitation behavior enhancement of the chimp optimization algorithm. Six canonical spiral shapes and two new hybrid spirals (HSS1 and HSS2) were embedded into the exploitation phase of ChOA to better mimic the realistic spiral hunting behavior of chimps and to address slow or premature convergence in the original algorithm.

Extensive experiments on 23 classical functions, 20 CEC 2005 benchmarks, 10 CEC06-2019 100-Digit Challenge problems, and 12 CEC 2020 constrained engineering problems showed that SEB-ChOAs, especially the HSS1 variant, consistently provide improved convergence and solution quality. Statistical tests confirmed significant performance gains over PSO, GA, SMA, MPA, ALO, HGSO, and the original ChOA, while remaining competitive with strong benchmark optimizers such as jDE100 and DISHchain1e−12.

Future work includes exploring alternative mutation mechanisms or acceleration parameters within the spiral framework, extending SEB-ChOA to binary and multi-objective settings, and applying the algorithm to feature selection and hyper-parameter optimization of classifiers such as support vector machines and multilayer perceptions.

### Cite as:

algorithm using spiral exploitation behavior. Neural Computing and Applications, 36(9), 4763-4786.

https://doi.org/10.1007/s00521-023-09236-y

https://link.springer.com/article/10.1007/s00521-023-09236-y